\documentclass[letterpaper, 10 pt, conference]{ieeeconf}
\usepackage{amsmath,amsfonts}
\usepackage{array}
\usepackage{enumerate}
\usepackage[utf8]{inputenc}
\usepackage[T1]{fontenc}
\usepackage{amsfonts}
\usepackage{amssymb}
\IEEEoverridecommandlockouts 
\usepackage{amsmath}  
\usepackage{breqn}
\usepackage[caption=false]{subfig}
\usepackage{textcomp}
\usepackage{stfloats}
\usepackage{url}
\usepackage{verbatim}
\usepackage{graphicx}
\usepackage{cite}
\usepackage[linesnumbered, ruled]{algorithm2e} 
\usepackage{hyperref}
\usepackage{diagbox}
\usepackage{amsfonts}
\usepackage{makecell}
\usepackage{adjustbox}
\usepackage{float}
\usepackage{color}
\usepackage{booktabs}

\title{\LARGE \bf
Multi-Agent LLM Actor-Critic Framework for Social Robot Navigation
}

\author{Weizheng Wang$^{}$, Ike Obi$^{}$, and Byung-Cheol Min$^{}$
\thanks{SMART Laboratory, Department of Computer and Information Technology, Purdue University, West Lafayette, IN, USA. {\tt\small{[wang5716,obii,minb]@purdue.edu}.}}}

\begin{document}

\maketitle

\begin{abstract} Recent advances in robotics and large language models (LLMs) have sparked growing interest in human-robot collaboration and embodied intelligence. To enable the broader deployment of robots in human-populated environments, socially-aware robot navigation (SAN) has become a key research area. While deep reinforcement learning approaches that integrate human-robot interaction (HRI) with path planning have demonstrated strong benchmark performance, they often struggle to adapt to new scenarios and environments. LLMs offer a promising avenue for zero-shot navigation through commonsense inference. However, most existing LLM-based frameworks rely on centralized decision-making, lack robust verification mechanisms, and face inconsistencies in translating macro-actions into precise low-level control signals. To address these challenges, we propose SAMALM, a decentralized multi-agent LLM actor-critic framework for multi-robot social navigation. In this framework, a set of parallel LLM actors, each reflecting distinct robot personalities or configurations, directly generate control signals. These actions undergo a two-tier verification process via a global critic that evaluates group-level behaviors and individual critics that assess each robot’s context. An entropy-based score fusion mechanism further enhances self-verification and re-query, improving both robustness and coordination. Experimental results confirm that SAMALM effectively balances local autonomy with global oversight, yielding socially compliant behaviors and strong adaptability across diverse multi-robot scenarios. More details and videos about this work are available at: \url{https://sites.google.com/view/SAMALM}.
\end{abstract}



\section{Introduction}

Current advancement in robotics, exemplified by Tesla's Optimus and Boston Dynamics' Spot, are driving extensive research into human-robot cooperation and embodied intelligence \cite{HRI-tro-1, HRI-tro-2, HRI-ijrr-1, HRI-ijrr-2, humanoid}. To facilitate the seamless integration of robotics and artificial intelligence into daily human environments, research on socially-aware robot navigation (SAN) is crucial for broader robotic deployment. Recent SAN studies \cite{navidiff, hypersamarl, san-1, san-2, san-3} highlight the importance of integrating both human-robot interaction (HRI) and robot-robot interaction (RRI) into social robot path planners, as these interactions are inherently dynamic and heterogeneous.

To enable large-scale robotic applications, it is essential for robot agents to exhibit human-level cognitive abilities and adhere with social norms. Currently, deep reinforcement learning (DRL)-based approaches that integrate HRI inference with path planning \cite{navidiff, wang2023navistar, liu2023intention, CADRL} demonstrate strong benchmark performance during iterative DRL training. However, these pre-trained policies are often constrained to their original training datasets and environmental configurations, limiting their adaptability to new environments. 

To address this limitation, large language model (LLM)-powered executors can be employed to navigate unfamiliar spaces in a zero-shot manner, leveraging their powerful commonsense inference capabilities. For instance, \cite{shah2023lm} introduced multi-modal LLMs to infer contextual semantic correlations for generating macro-actions (destinations) in zero-shot navigation. Additionally, \cite{kannan2024smart, chen2024llm} extended this concept to multi-robot scenarios by utilizing LLM-based path planners as task allocators. However, these methods rely on centralized decision-making systems that fail to account for the unique configurations and preferences of individual robots. Furthermore, they lack the ability to directly control robots or multi-robot systems via low-level controller signals.

Most existing zero-shot LLM-based approaches face several challenges: (1) the centralized single-LLM execution paradigm struggles to account for the unique characteristics and properties of different robot types; (2) low-level robot control signals (e.g., [$v_x, v_y$]) are typically derived from macro-actions, creating a disconnect between LLM inference and the robot controller, which can lead to reasoning instability and inconsistency; and (3) the absence of a verification step undermines the robustness and cooperativeness of multi-robot systems, increasing the risk of LLM-induced hallucinations or numerical errors.

\begin{figure}[!t]
\centering
\includegraphics[width=1\columnwidth]{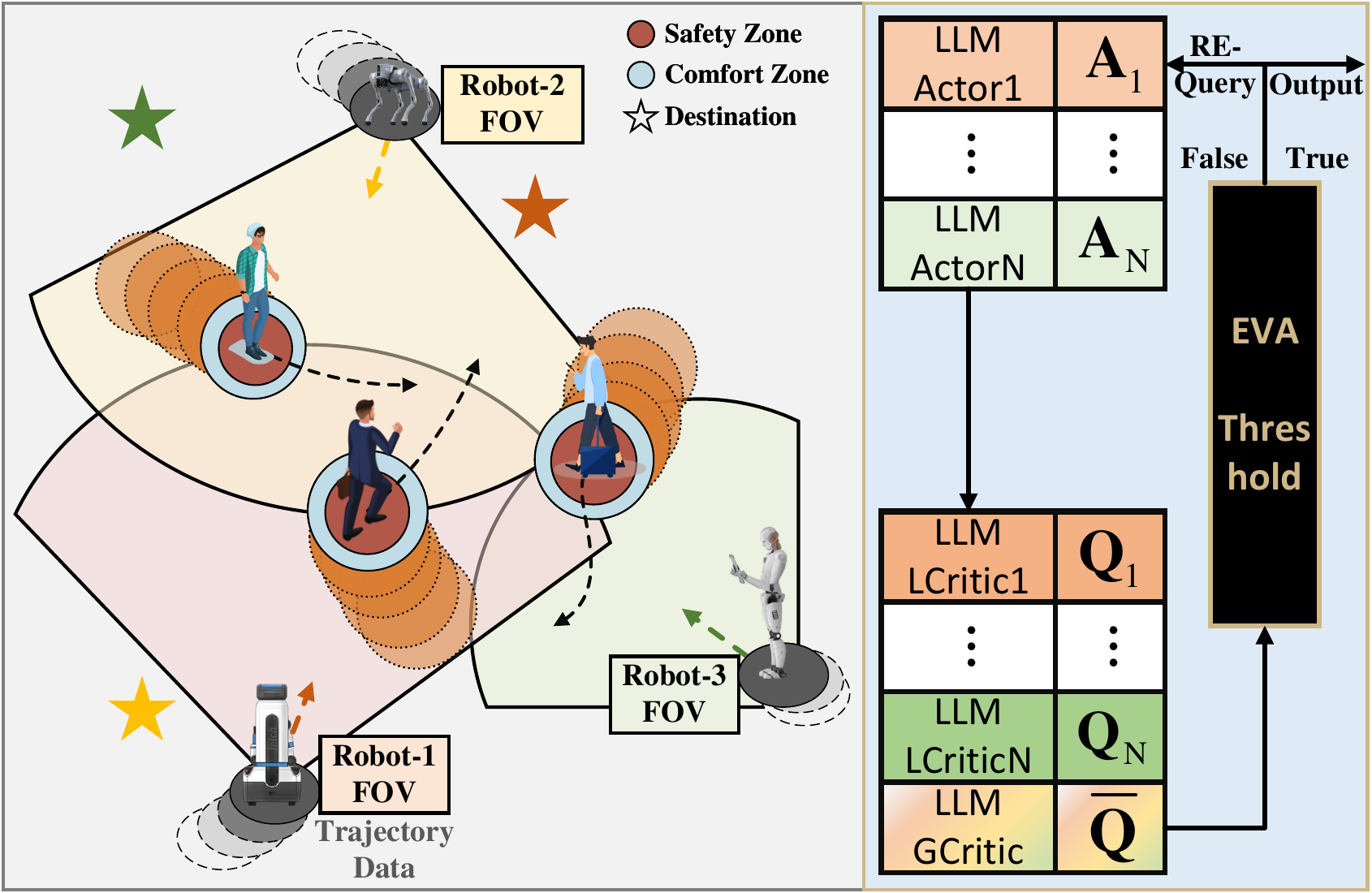}
\vspace{-10pt}
\caption{SAMALM procedure in a multi-robot social navigation scenario: The decentralized decision-making system supports self-verification and re-query, consisting of multiple parallel LLM-actors, evaluated by both the individual LLM-critic and a global LLM-critic.}
\vspace{-20pt}
\label{fig:F1}
\end{figure}

Inspired by the successes of multi-LLM coordination in multi-agent paradigm \cite{stanfordtown, chatdev}, this paper presents a decentralized multi-agent LLM actor-critic framework that supports self-verification and     re-querying to address the aforementioned challenges in multi-robot SAN tasks, as shown in Fig.~\ref{fig:F1}. Specifically, a set of individual LLM-based actors is designed to reflect the unique personalities or characteristics of different robots, enabling them to generate direct control actions. These actions are then evaluated and refined through a two-tier verification process until satisfactory results are achieved: a global critic assesses group-level behaviors of the multi-robot system, while individual critics evaluate each robot's actions in context. The main contributions of this paper can be summarized as follows: 
\begin{itemize}
  \item We propose a novel \textbf{S}ocially-\textbf{A}ware \textbf{M}ulti-\textbf{A}gent actor-critic L\textbf{LM} framework, called SAMALM, to address multi-robot social navigation tasks.
  \item SAMALM develops the multi-LLM world model for understanding environmental dynamics, where  the world model constructs personal knowledge of the task scenario based on each robot’s interactions with the environment. Additionally, SAMALM employs a multi-LLM agent system with the personalized prompt engineering to represent distinct robot preferences, allowing for asynchronous queries from each LLM actor or critic.
  \item SAMALM leverages a multi-LLM actor-critic framework for self-verification and enhanced group cooperation, incorporating a global LLM critic and a set of local LLM critics with an entropy-based score fusion mechanism.
  \item SAMALM demonstrates promising socially compliant behaviors across various experiments.
\end{itemize}


\section{Background}


\subsection{Multi-Agent LLM Framework for Robotics}

The advancements in LLMs have dramatically broadened the scope of artificial intelligence, revolutionizing natural language processing by delivering unprecedented capabilities in language generation and semantic understanding. Recently, breakthroughs in LLMs have spurred innovative research in robotics and embodied intelligence \cite{shah2023lm, ichter2022do}, where LLMs' commonsense inference abilities are leveraged for contextual environmental understanding and execution generation, even in zero-shot or few-shot settings. Moreover, integrating multiple LLMs allows researchers to leverage their diverse strengths and mitigate individual limitations, as demonstrated in applications such as multi-agent communication \cite{stanfordtown} and cooperative software development \cite{chatdev}. By orchestrating several specialized models within a unified framework, multi‐LLM systems strive to enhance performance on complex tasks, improve reliability, and facilitate more nuanced reasoning.

The integration of LLMs into multi-robot systems has unlocked transformative opportunities in communication, task planning, social navigation, and human-robot interaction. LLMs excel at interpreting intricate instructions, reasoning through multifaceted tasks, and facilitating natural communication. When deployed within multi-robot frameworks, employing multiple LLMs each specialized for distinct robots such as unmanned aerial vehicles, humanoid robots, or mobile robots enables the entire system to heterogeneously achieve robust coordination, adaptive decision-making, and enhanced autonomy in complex environments. This multi-LLM approach not only leverages the unique strengths of individual models but also creates a synergistic network capable of addressing the diverse challenges encountered in dynamic scenarios.

Hence, the success of the multi-agent LLM paradigm has also motivated the development of LLM-powered multi-robot systems. For example, \cite{kannan2024smart} introduced SMART-LLM, a framework that decomposes high-level natural language instructions into detailed sub-tasks while orchestrating coalition formation and task allocation among robots. Similarly, \cite{obata2024lip} leverages linear programming and dependency graphs to capture the latent correlations between task steps in LLM-based multi-robot cooperation. However, these systems are limited by their reliance on centralized LLM planners, which struggle to adequately represent the heterogeneous attributes of individual robots. Alternatively, \cite{mandi2024roco} and \cite{wu2024autogen} designed the multi-robot system planner with respect to the decentralized multiple LLMs. Herein, SAMALM also explicitly consider each robot's unique types and properties within a decentralized multi-LLM actor-critic framework, enabling tailored coordination and adaptive task execution that better captures the heterogeneity inherent in the multi-robot SAN task.

\subsection{Large Language Models for Social Navigation} 

As a fundamental topic in robotics, socially-aware navigation has recently benefited from the advent of LLMs as promising tools for robust task execution. While state-of-the-art DRL-based approaches such as GST \cite{liu2023intention}, NaviSTAR \cite{wang2023navistar}, and NaviDIFF \cite{navidiff} have set performance benchmarks, these pre-trained DRL-based strategies often struggle when transferred to unfamiliar environments. To adapt environmental generalization, LLM-based social planners have been developed for zero-shot or few-shot scenarios. Benefiting from LLM's exceptional commonsense understanding, LLM-based social planners also demonstrate anticipated navigation capabilities. For example, \cite{sr-llm2} introduced a multi-task social navigation planner that hybridizes LLMs with DRL to enhance policy transferability. Moreover, recent advances in visual language models have shown particular promise in supporting social navigation tasks \cite{sr-llm1}, \cite{sr-llm3}. These models parse perception images from robot cameras to extract latent cues for HRI inference and socially-aware path planning.

Furthermore, SAN tasks have been extended to multi-robot scenarios \cite{wang2023multi}. In these settings, these planners have to consider not only latent HRI features but also RRI and system-level cooperative behavior execution among the multi-robot system, throughout the navigation process. \cite{wang2023multi} leveraged spatial-temporal transformer and multi-agent proximal policy optimization algorithm to address the multi-robot SAN task within the Dec-POSMDP (decentralized partially observable semi-Markov decision process). \cite{dong2024multi, escudie2024} developed neural network architectures based on gated recurrent unit and graph attention network to estimate latent HRI and RRI for coordinated path planning. These models are further fine-tuned using multi-agent reinforcement learning as well. More recently, \cite{hypersamarl} unified multi-robot task allocation and social navigation by leveraging a hypergraph diffusion mechanism that captures high-order correlations among robots, humans, and points-of-interest. This innovative approach facilitates a comprehensive understanding of the intricate dependencies within the system, thereby enhancing coordinated performance across complex environments.

Apart from that, LLM-based path planners have also been adapted to this scenario. Despite existing LLM-based path planners \cite{chen2024llm, kannan2024smart} have been developed for cooperative navigation, the specific paradigm of multi-robot socially-aware navigation remains largely unexplored in current research. Thus, to the best of our knowledge, we are the first to implement a multi-LLM framework, termed SAMALM for multi-robot SAN tasks. SAMALM addresses dynamic semantic inference in both HRI and HHI to enhance coordinated navigation behavior representation. Moreover, the framework leverages a self-verification and re-query procedure via a multi-LLM actor-critic approach, refining social robots' executions based on evaluations from both group-level and agent-level perspectives.

\vspace{-5pt}
\section{Preliminary}

\begin{figure*}[!t]
\centering
\includegraphics[width=0.95\linewidth]{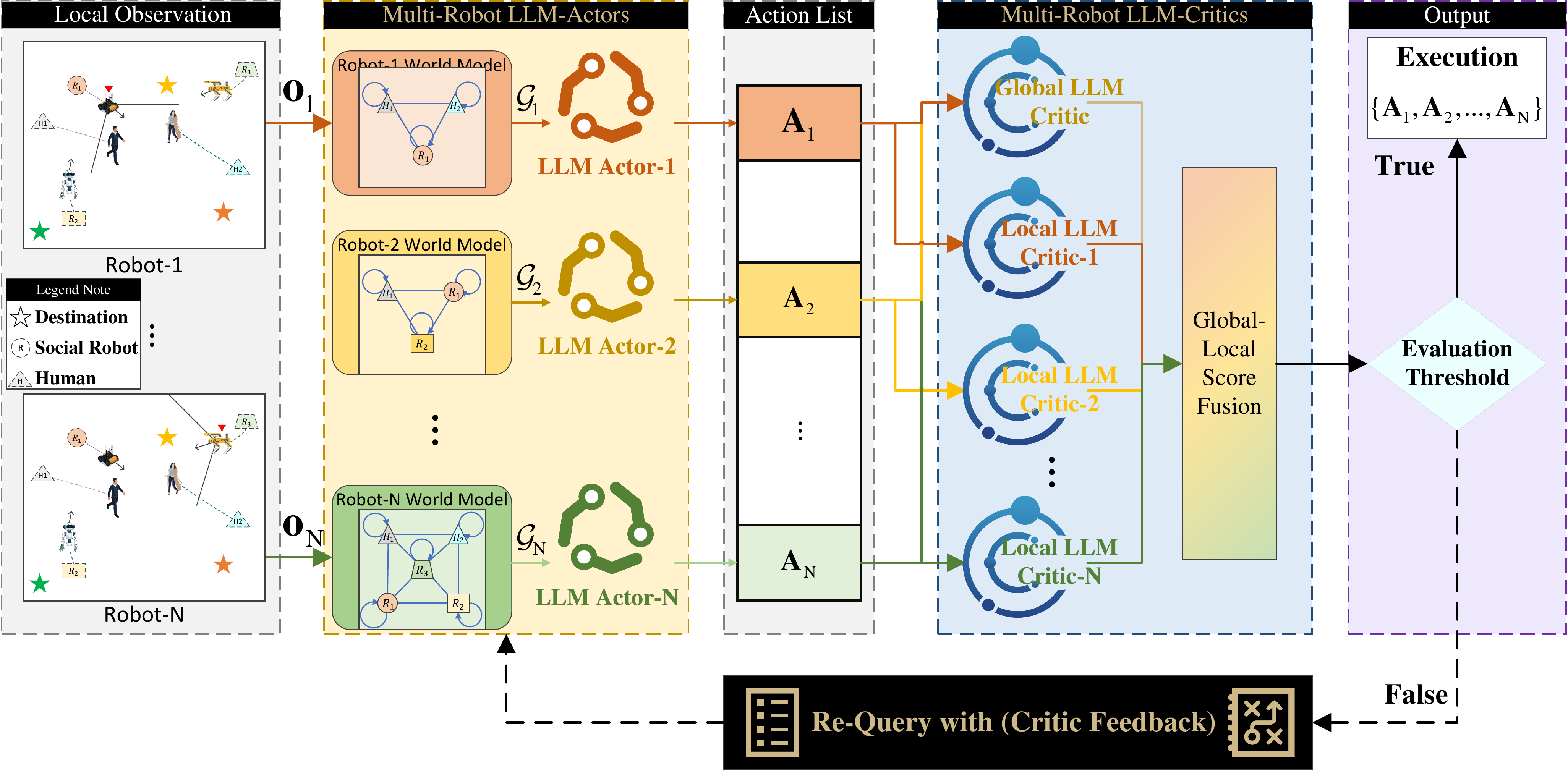}
\vspace{-10pt}
\caption{SAMALM architecture: SAMALM is a decentralized multi-agent LLM actor-critic framework designed for multi-robot social navigation. In SAMALM, a set of LLM-actors generates low-level control signals for the robots, respectively. These generated actions are then evaluated by relative LLM-critics from both the team-level and agent-level perspective, which either confirms the actions or prompts a re-query with critic feedback. Once the actions pass the evaluation threshold, they are executed by the system’s executors in the multi-robot environment.}
\vspace{-15pt}
\label{fig:F2}
\end{figure*}

According to \cite{navidiff, hypersamarl, CADRL}, we reformulate the multi-robot social navigation task as a textual decentralized partially observable Markov decision process (Dec-POMDP) $\langle \mathbf{\hat{S}}, \mathbf{\hat{A}}, \mathbf{\hat{O}},\mathbf{\hat{P}}, \mathbf {\hat{S}}_0, n \rangle$.
Here, the joint observations, denoted as $ \hat{o}= [o_1, o_2, \cdots, o_n] \in \mathbf{\hat{O}}$ are transformed into a multi-robot world model representation that serves as input for a set of LLM-actors. Subsequently, individual actions $ \hat{A} = [A_1, A_2, \cdots, A_n] \in \mathbf{\hat{A}}, ~A = [v_x, v_y]$ are generated as multi-robots' executions. In the formulation, $n$ denotes the number of social robots, and $\mathbf{\hat{S}}_0$ presents the initialized state distribution. In contrast to traditional MDP, we convert the observation and action pairs into textual representations during the interaction between the robots and their environment. This textualization via world model is essential for the effective deployment of our LLM-based actor-critic framework.

At each timestep, the system collects local observations from the environment to generate actions. These actions interact with the environment, updating the system state $\mathbf{\hat{S}}$ corresponding to the system’s translation function $\mathbf{\hat{P}}$ and the detection of conditional events such as arrivals or collisions. Notably, human actions are driven externally by individual motion policies that is also integrated into the simulation.

Furthermore, due to the deployment of LLMs, SAMALM processes environmental feedback by generating both a numerical score and descriptive text through a set of critics. This design establishes a self-verification mechanism: if an action fails to meet a predefined evaluation threshold, it is re-queried along with the associated critic reasoning text. This approach removes the need for a conventional reward function in the original MDP configuration, while still preserving the environmental translation function based on uniform motion kinematics.

\section{Methodology}

\begin{figure*}[!t]
\centering
\includegraphics[width=0.95\linewidth]{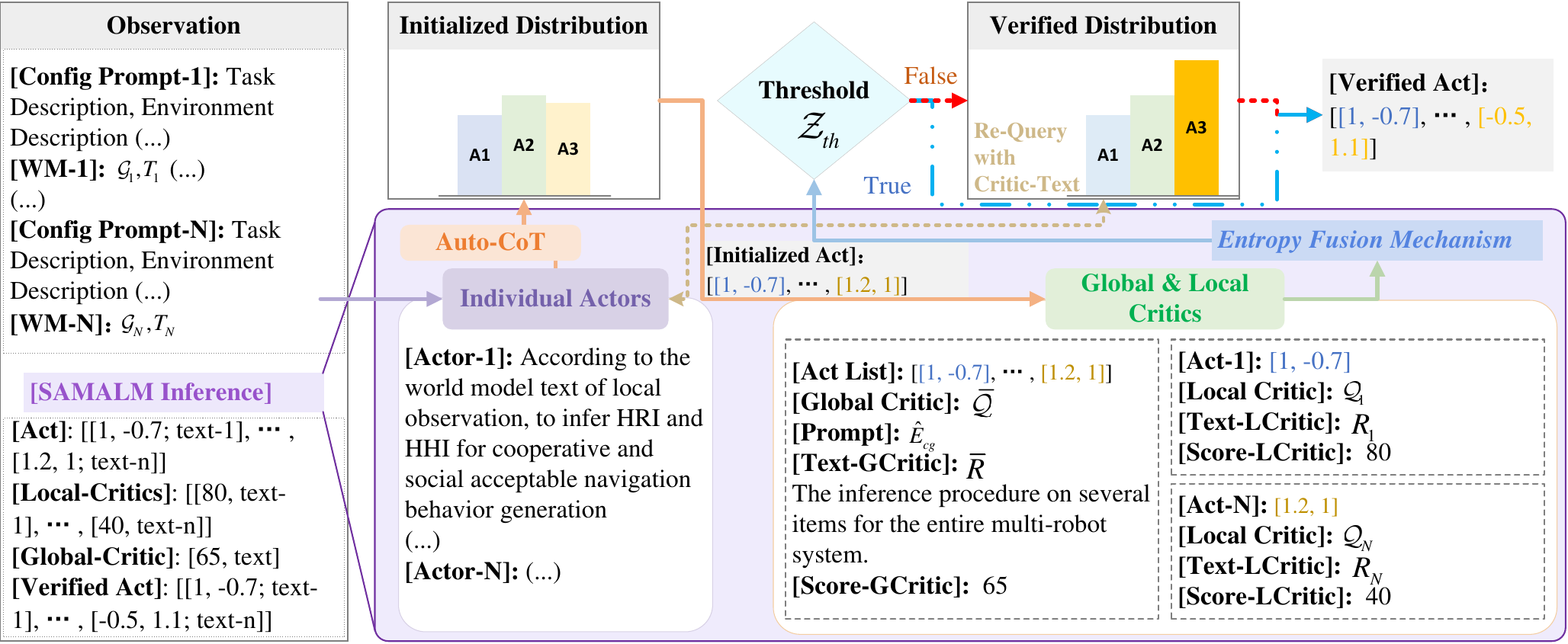}
\vspace{-10pt}
\caption{Multi-Agent LLM Actor-Critic Framework: SAMALM facilitates multi-robot social navigation using a set of parallel LLM actors that extract semantic correlations from local world model observations and work in tandem with both global and local critics. The global critic assesses multi-robot behaviors by considering both inter-group and intra-group dynamics, while local critics evaluate individual actions based on long-term and short-term factors. Ultimately, the global and local critic scores are integrated via an entropy-based fusion mechanism that accounts for the level of disagreement among the critics, enabling self-verification and re-query with critic feedback.}
\vspace{-15pt}
\label{fig:F4}
\end{figure*}

SAMALM presents a decentralized multi-robot cooperative social navigation planner that combines an actor executor and a critic scorer framework for each individual robot, all powered by LLMs. In contrast to previous works \cite{kannan2024smart, chen2024llm}, SAMALM not only enables each robot’s LLM actor to generate actions tailored to its unique preferences, but also employs a global LLM critic to verify team-level coordination and local LLM critics to maintain individual rationality. In our multi-robot configuration, a bi-directional communication message chain facilitates information sharing, directing whole team data to the leader robot for global critic's evaluation. SAMALM focuses on representing cooperative behaviors through an LLM-powered multi-robot framework, supporting self-verification and re-query capabilities within a multi-agent actor-critic architecture. Our approach is depicted in Fig.\ref{fig:F2}, with further details on the multi-robot world model in Fig.\ref{fig:F3} and the multi-agent actor-critic framework in Fig.~\ref{fig:F4}.

\subsection{Multi-Agent World Model}



\begin{figure}[!t]
\centering
\includegraphics[width=1\columnwidth]{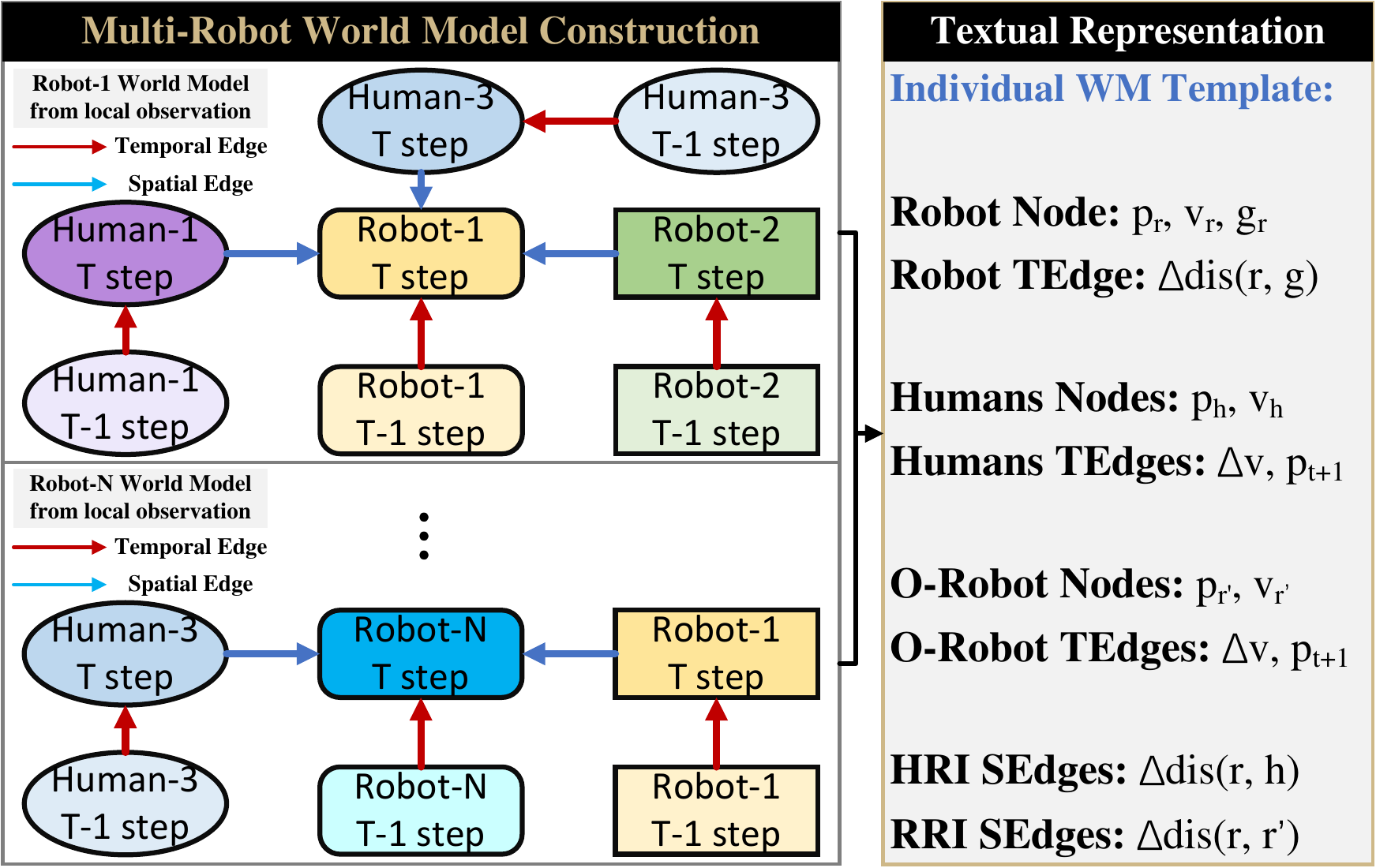}
\vspace{-10pt}
\caption{An illustration of multi-robot world model construction.}
\vspace{-10pt}
\label{fig:F3}
\end{figure}

Humans maintain a dynamic internal thought paradigm, a richly detailed and continuously updated mental representation of their surroundings that enables them to simulate potential actions and accurately predict their outcomes. This sophisticated cognitive framework underpins not only basic functions like motor control and sensory perception but also higher-level abilities such as mental imagery, logical inference, and strategic decision making. Inspired by recent research on world models applied to mathematical problem \cite{wm1} and robot navigation \cite{wm3, wm-nav-2}, we develop a spatio-temporal graph structural multi-robot world model for textually representing human-robot interactions (HRI) in LLM-based multi-robot SAN tasks, as shown in Fig.~\ref{fig:F3}.

Formally, the multi-agent world model $\{ \mathcal{G}_1, \cdots, \mathcal{G}_n \}$ is constructed to facilitate the understanding of HRI and RRI for each robot by capturing both environmental dynamics. It describes the interactive process with spatial-temporal correlations modeled as a graph denote the vertices or nodes corresponding to robots and humans, respectively, while represent the spatial and temporal edges that capture their interactions. Notably, SAMALM constructs a dedicated spatial-temporal graph for each individual robot, derived from its unique local observations, thereby ensuring tailored and contextually representations as follows:
\begin{equation}
    \{ T_1, \cdots, T_n\} = \{ \mathcal{G}_1(\mathcal{V}_1, \mathcal{E}_1 ~|~ o_1), \cdots, \mathcal{G}_n(\mathcal{V}_n, \mathcal{E}_n ~|~ o_n) \}
\end{equation}
where $T_i$ denotes the textual world model representation.

In details, the world model nodes of $i$-th robot, denoted as $\mathcal{V}_i = \{ \mathcal{V}_r, \mathcal{V}_R, \mathcal{V}_H \}$, consist of following components: self-node $\mathcal{V}_r = [p_r, v_r, g_r]$ contains this robot's own location, velocity, and goal information; observed humans and other robots nodes $\mathcal{V}_R, \mathcal{V}_H = [p, v]$ provide the position and velocity data with respect to robot observation $o_i$. As shown in Fig.~\ref{fig:F3}, the spatial relationships among these nodes are modeled using HRI and RRI spatial edges, which describe both the trend and value of the relative distance dynamics between the robot and the observed entities at the current timestep. Furthermore, the temporal edge associated with the robot’s self-node represents the relative distance to its destination. For the observed human and robot nodes, the temporal edges capture both the acceleration (i.e., the change in speed) and next timestep's position estimation, based on uniform linear motion kinematics using the current velocity.

\subsection{Multi-Agent LLM Actors}



Figure~\ref{fig:F4} illustrates that we leverage the LLM directly as the robot’s policy network, utilizing its strong commonsense inference capabilities to understand latent environmental dynamics for robot action generation. Unlike previous approaches \cite{shah2023lm, chen2024llm}, SAMALM’s LLM-actors produce direct control commands $A = [v_x, v_y]$, rather than macro actions or instructions, thereby preserving inference consistency and coherence.

Based on \cite{actor, chen2024llm}, the prompt engineering for LLM-actors involves two key components: the task configuration textual description $\bar{T}$, and the environmental dynamics textual description $T_i$, derived from the world model. The information in $\bar{T}$ which includes the overall task and data descriptions, is commonly shared to all actors. In contrast, $T_i$ captures the specific observations and HRI semantic details pertinent to each actor. Eventually, the chain chaining inference technology, Auto-CoT \cite{autocot} is employed to enhance actors' reasoning ability as follows:
\begin{equation}
    \mathcal{P}_{i}( a_i ~| ~T_i, \bar{T}, E_i ) = \mathcal{P}_{i}(a_i ~|~ I_{[s]}) \times \mathcal{P}_{i}(I_{[s]} ~|~ T_i,\bar{T}, E_i) 
\end{equation}
where $E_i$ is the prompt engineering with $i$-th robot's preference, and $I_{[s]}$ denotes the total $s$-step reasoning chain that is generated by Auto-CoT \cite{autocot}.

Inspired by \cite{wu2024autogen,jiang2023evaluating}, SAMALM integrates robot-specific preferences into its prompt engineering framework to accommodate the diverse attributes of different robotic platforms. This integration focuses on two key aspects: the robot's speed preference and its designated socially acceptable distance. For example, robotic dogs and humanoid robots have different speed configurations and distinct notions of personal space, which SAMALM addresses by tailoring the prompt parameters accordingly.

\subsection{Multi-Agent LLM Critics}

Once the robot actions are generated by LLM-actors, the self-verification mechanism is utilized to evaluate and grade such actions via a set of LLM-critics with a rule-based checklist prompt $E_{c(\cdot)}$. As shown in Fig.~\ref{fig:F2} and Fig.~\ref{fig:F4}, SAMALM is composed by both individual LLM-critics $\mathcal{Q}_i(o_i, A_i, E_{ci})$ and a global LLM-critic $\mathcal{\bar Q}_g (\hat{o}, \hat{A}, \hat{E}_{cg})$. According to \cite{critic}, SAMALM motivates LLM-critics to verify robot action with respect to external curial checklist and current observation condition. Considering both preferred speed and type of robot personality. In details, the prompt engineering of each local LLM-critic is designed to evaluate robot action via both short-term and long-term penalty $\gamma(l)$. In $\mathcal{Q}_i$'s short-term evaluation, local critic is encouraged to check the relative distance among local robot and neighbor humans with respect to current situation and next timestep's prediction. Meanwhile, the long-term factors check the appearance of high risk area (more than $n_{th}$ humans will be involved within the uncomfortable distance area of the $i$-th robot) on next $k$-timestep that when as follows:
\begin{equation}
	\gamma_i(l)=
	\begin{cases}
	    -5 \times n_h^t,& if~dis(r,h) < dis_{s} \\
	    -10 \times n_h^t,& if~dis(r,h) < dis_{c} + 0.1 \\
	    -5 \times (n_h^{t+k}-3), & if ~ n_h^{t+k}>n_{th}
    \end{cases}
\end{equation}
where $n_h^t$ denotes the humans who are involved within uncomfortable distance from local robot on $t$-timestep. $dis_s = \rho_{pref} + \rho_r + \rho_h$ present the social distance, and $\rho_{pref}$ is the social acceptable distance of robot personality configuration, and $dis_c = \rho_r + \rho_h$ is the collision distance. The local critic score is defined as $\mathcal{Q}_i = 100 - \sum(\gamma_i(l)) \in (-\infty, 100]$.

Similarly, the global critic applies penalties that account for both intra-group and inter-group evaluation factors. The intra-group factor emphasizes cooperative navigation and collision detection among robots, while the inter-group factor focuses on interactions between the multi-robot system and humans, as well as the overall effectiveness of the navigation task. Specifically, the global critic penalty $\gamma(g)$ is defined as follows:
\begin{equation}
	\gamma(g)=
	\begin{cases}
	    -10 \times n_r^t,& if~dis(r,r') < dis_{c} + 0.1 \\
	    -1.25 \times (t-t_{m}),& if~t > 30 \\
	    -15 \times (\#R^{t+k}_{hr}), & if ~ \#R^{t+k}_{hr} > \#R/2
    \end{cases}
\end{equation}
where $n_r^t$ denotes the other robots who are involved within collision distance from local robot on $t$-timestep, and $t_m$ is the hyperparameter of the average navigation time that refer to the robot types, and $\#R^{t+k}_{hr}$ presents the number of robots who are surrounding within more than $n_{th}$ humans' uncomfortable distance area, and $\#R$ is the number of robots. Lastly, the global critic score is defined as $\mathcal{\bar Q} = 100 - \sum(\gamma(g)) \in (-\infty, 100]$.

\begin{algorithm}[h]
 Initialization: Initialize the joint state $\hat{s}_o \in \mathcal{S}_0; ~ \text{LLM-Actors}~ \{\mathcal{P}_1,\cdots, \mathcal{P}_N\}; ~\text{LLM-Critics} ~\{\mathcal{Q}_1 , \cdots,$ 
 $\mathcal{Q}_N, \mathcal{\hat Q} \},~\text{and other hyperparameters}$\;
Reset the environment and data buffer $\mathcal{D}$\;
\For {$step\leq step_{max}$}{
	 Update robots' observations $\hat o_t$\;  

      Construct world models $\{T_1, \cdots, T_N\}$ by Eq.(1)\; 

      Sample actions from LLM-Actors $A_i \sim \mathcal{P}_i(T_i) $, to update joint action $\{A_1, \cdots, A_N\}$ by Eq.(2)\;

      Query scores from LLM-Critics $\{\mathcal{\hat Q}, \mathcal{\bar{Q}}, \hat{R}, \bar{R}\}$\;

      Calculate critic fusion score $\mathcal{Z}$ by Eq.(8)\;

      \While{$\mathcal{Z} < \mathcal{Z}_{th}$}{
      Re-query $\forall A_i \in \hat{A}, ~ \mathcal{Q}_i < \mathcal{Z}: A^{'}_{i} = \mathcal{P}_i(T_i, \mathcal{D}_i)$\;
      Generate new verification $\{\mathcal{Q}_i^{'}, \mathcal{\bar{Q}}^{'}, {R}_i^{'}, \bar{R}^{'}\}$\;
      Update the data buffer $\mathcal{D}_i = [{R}^{'}_i, \bar{R}^{'}]$\;
      Calculate and update the fusion score $\mathcal{Z}^{'} = f(\mathcal{Q}_i^{'}, \mathcal{\bar{Q}}^{'}),$ $\mathcal{Z} = \mathcal{Z}^{'}$ by Eq.(8) \;
      }
      Generate the execution $\hat{A}$\;
      Environment transformation $\hat{s}_{t+1} = \mathbf{\hat{P}}(\hat{s}_t, \hat{A}_t)$\;
}      
\caption{Socially-Aware Multi-Agent LLM}
\label{alrorithm1}
\end{algorithm}
\setlength{\textfloatsep}{0.15cm}

\subsection{Entropy Fusion Mechanism}

Herein, we propose an entropy-based critic scores fusion mechanism, which adaptively integrate both local scores and global score with respect to the disagreement-level weighting scheme. Firstly, each local critic score $\mathcal{Q}_i$ is normalized to produce the confidence distribution $\mathcal{C}_i$ of the sum as follows:
\begin{equation}
    \mathcal{C}_i = \frac{\mathcal{Q}_i}{\sum_{j=1}^{N} \mathcal{Q}_j}
\end{equation}

Subsequently, the entropy of each local score is defined to estimate the disagreement-level among local critic evaluations as follows:
\begin{equation}
    \mathcal{H}_i = - \kappa {\sum}_{i=1}^{N} \mathcal{C}_i ~ \log(\mathcal{C}_i)
\end{equation}
\noindent where $\kappa$ is the entropy hyperparameter, $\log$ is the logarithm with the constant $e$ base.

Hence, the system local critic score entropy can be obtained as $\mathcal{H} = \sum_{i=1}^{N} \mathcal{H}_i$ from above equation. Furthermore, the disagreement weight $\omega$ between global critic score and local critic scores is defined as follows:
\begin{equation}
    \omega = \frac{\mathcal{H}}{\mathcal{H}_{max}} = \frac{\mathcal{H}}{ \log(1/N)}
\end{equation}
\noindent where $\mathcal{H}_{max}$ denotes the maximum entropy of multi-robot system local critic scores, implying the lowest evaluation disagreement among local critics.

Eventually, the critic fusion score $\mathcal{Z}$ is leveraged by the entropy-based weighted local critic scores and global critic score as follows:
\begin{equation}
    \mathcal{Z} = \omega (\frac{1}{N}{\sum}^{N}_{i=1} \mathcal{Q}_i) + (1-\omega) \mathcal{\bar{Q}}_g
\end{equation}

Notably, the critic fusion score $\mathcal{Z}$ implies an evaluation tendency that favors local critic scores when they are more centralized, indicating a high degree of consensus. Conversely, when local scores are ambiguous and disordered, the global score becomes more influential. After the estimation of critic fusion score, the evaluation threshold is set as a hyperparameter $\mathcal{Z}_{th}$ to determine whether to re-query or proceed with execution. If the robot actions meet the threshold as $\mathcal{Z} \geq \mathcal{Z}_{th}$, they are transmitted as controller signals to drive the robot. Otherwise, any local action with a critic score lower than the fusion score $\mathcal{Z}$ is updated by the corresponding actor, incorporating evaluation reasoning records from the relative local critic $R_i$ and the global critic $\bar{R}$. For example, the critic reasoning record (e.g. a critic might indicate that the action would bring the robot too close to human-7) is shared to actor for the enhancement, avoiding the same issue.

\vspace{-5 pt}
\section{Experiments}

In this section, we evaluate the performance of our proposed SAMALM framework in multi-robot social navigation tasks. The experiments are designed to assess not only the multi-robot navigation success but also the degree of social compliance exhibited by the robots. To this end, we compare our complete models against several baselines and ablation variants, under a simulated environment that replicates realistic human-populated scenarios.

\begin{figure}[!t]
\centering
\includegraphics[width=0.65\linewidth]{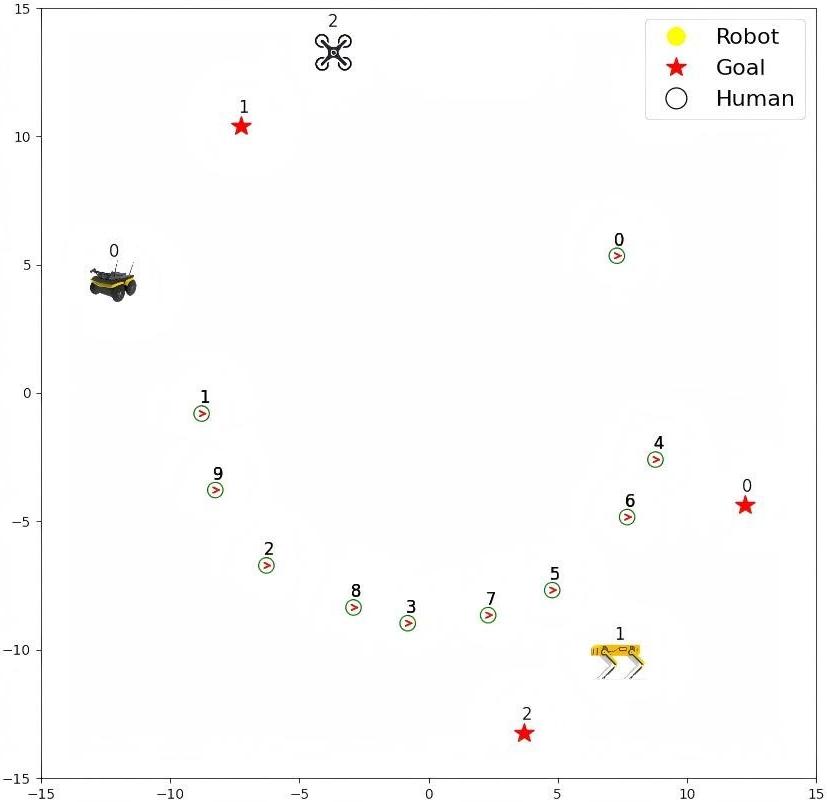}
\vspace{-3pt}
\caption{The illustration of heterogeneous multi-robot social navigation simulation scenario, where 3 robots are navigating by individual LLM-actor to distinct destinations across the crowd with 10 humans.}
\vspace{-5pt}
\label{fig:box}
\end{figure}

\subsection{Experimental Setup} 
As shown in Fig.~\ref{fig:box}, our experiments are conducted in a simulation environment consistent with previous works \cite{hypersamarl, wang2023multi}, where each robot is endowed with a $F\in(0, 2\pi]$ degree field-of-view (FOV) to ensure locally situational awareness. The multi-robot system is controlled by a decentralized LLM-powered decision-making paradigm. In this setup, individual robots operate autonomously, making decisions based on local observation data and communicating directly with nearest robot along the message chain \cite{parasuraman2018multipoint}, rather than relying on a centralized controller. This decentralized approach enables each robot to process information, plan actions, and coordinate with its peers, leading to emergent collective behavior that reflects both group-level intelligence and individual intrinsic properties such as preferred speed and acceptable social distance. Particularly, the communication chain of multi-robot system is employed to facilitate connection among agents, enabling the system to leverage both local and global information during decision-making. 


In our setup, social robots are tasked with cooperatively navigating toward distinct destinations while maintaining appropriate social distances from pedestrians. We set the configuration of three typical robot types with respect to mobile robot (preferred velocity = 1.25 m/s; social distance = 0.45 m), robot dog (preferred velocity = 1 m/s; social distance = 0.3 m), and drone (preferred velocity = 1.5 m/s; social distance = 0.85 m). Additionally, these three robot types are randomly assigned for 5-robots 10-humans scenario. Human pedestrians are simulated using a personalized policy \cite{ORCA}, which allows them to change velocities and adjust their goals randomly, reflecting realistic behavior. Importantly, sensitive information regarding pedestrian policies and intended goals remains inaccessible to the robots.


\subsection{Baselines and Ablation Models} 

We evaluate three categories of models. First, we select several baseline models: LLaMA-8B, LLaMA-70B, LLaMA-405B \cite{llama}, GPT-3.5, GPT-4, and GPT-4o \cite{openai2023gpt4}. These baselines are employed within a centralized LLM decision-making framework and Auto-CoT inference procedure, where they convert aggregated observations from all robots directly into execution commands, without utilizing an actor-critic structure.

For the ablation studies, we construct two variants using LLaMA-405B and GPT-4o as the LLM backbones, but without the critic modules; these are denoted as Ablation-L and Ablation-G, respectively. In contrast, the full SAMALM model incorporates the critic structure to enable self-verification support. The zero-shot baselines directly employ large language models for planning without additional verification, while the ablation studies help quantify the contribution of the critic modules. In our full SAMALM models, the two-tier critic system (local and global) coupled with an entropy-based fusion mechanism ensures that initial action proposals are iteratively refined until they satisfy a predefined evaluation threshold.

\subsection{Evaluation Metrics} Two primary metrics are used to gauge performance within total 50 same testcases: (1) Success Rate (SR): This metric quantifies the percentage of episodes in which the robots successfully reach their designated goals without collisions or timeouts; (2) Social Score (SS): Derived from \cite{wang2023multi}, this score assesses the quality of robot behaviors with respect to social compliance. It is computed based on several factors, including path quality, adherence to personalized safe distances, and the frequency of human discomfort incidents.



\section{Experiments And Results}


\begin{table}[h]
\vspace{-5pt}
\caption{Simulation Experiment Results\label{tab:table1}}
\vspace{-5pt}
\centering
\begin{scriptsize}
\begin{tabular}{cccccccc}
\hline & \multicolumn{3}{c}{ Success Rate } & & \multicolumn{3}{c}{MR-SAN Social Score } \\
\cline { 2 - 4 } \cline { 6 - 8 } Methods & \multicolumn{3}{c}{ FOV\&Human\&Robot } & & \multicolumn{3}{c}{ FOV\&Human\&Robot } \\
& 90° & 360° & 360° & & 90°  & 360° & 360° \\
&5\&3 & 5\&3 & 10\&5 && 5\&3 & 5\&3 & 10\&5\\
\hline 
LLaMA-8B & $0$ & $0$ & $0$ & & $0$ & $0$ & $0$ \\
LLaMA-70B & $4$ & $4$ & $0$ & & $1$ & $0$ & $0$ \\
LLaMA-405B & $22$ & $14$ & $0$ & & $15$ & $8$ & $0$ \\
Ablation-L & $34$ & $30$ & $4$ & & $29$ & $21$ & $2$ \\
SAMALM-L (ours)   & $\mathbf{44}$ & $\mathbf{42}$ & $\mathbf{12}$ & & $\mathbf{31}$ & $\mathbf{31}$ & $\mathbf{5}$\\
\hline
GPT-3.5   & $0$ & $0$ & $0$ & & $0$ & $0$ & $0$\\
GPT-4   & $6$ & $6$ & $0$ & & $3$ & $2$ & $0$\\
GPT-4o & $32$ & $26$ & $2$ & & $27$ & $15$ & $0$ \\
Ablation-G & $56$ & $52$ & $18$ & & $47$ & $42$ & $15$ \\
SAMALM-G (ours)   & $\mathbf{68}$ & $\mathbf{70}$ & $\mathbf{30}$ & & $\mathbf{72}$ & $\mathbf{69}$ & $\mathbf{25}$\\
\hline
\end{tabular}
\end{scriptsize}
\vspace{-7pt}
\label{table:result}
\end{table}

\subsubsection{Baselines Result}


Table~\ref{table:result} summarizes our experimental findings and underscores the effectiveness of the SAMALM framework for multi-robot social navigation. Across various configurations, SAMALM consistently outperforms both traditional centralized baselines and its ablation variants. For example, in a 90° FOV scenario with 5 robots and 3 humans, SAMALM-L achieves an SR of 44 and an SS of 31—double the performance of the LLaMA-405B baseline (SR 22, SS 15). Similarly, SAMALM-G records an SR of 68 and an SS of 72, substantially surpassing the GPT-4o baseline.

In contrast, earlier LLMs such as LLaMA-8B and GPT-3.5 yield negligible performance in multi-robot SAN tasks, as evidenced by zero-averaged SR and SS. Their output files show that these models produce only fixed actions (e.g., [[0, 0], ..., [0, 0]]), and their trajectory histories reveal a clear inability to process multi-variable inputs and task requirements. Although larger models like LLaMA-70B and GPT-4 perform better than their smaller counterparts, they still struggle often making value errors and variable miscalculations—rendering their occasional successes largely fortuitous. Notably, LLaMA-405B and GPT-4o demonstrate promising reasoning abilities by accurately performing calculations and properly utilizing variables, resulting in average scores of (SR = 12, SS = 8) and (SR = 20, SS = 18), respectively.



\subsubsection{Ablation Study}

Table~\ref{table:result} demonstrates that the multi-LLM actor-critic framework substantially improves performance. For the LLaMA-based SAMALM, the averaged SR increases from 23 to 33 with a $30\%$ improvement, while the averaged SS rises from 17 to 22, representing a $23\%$ gain compared with the corresponding ablation models. Similarly, for the GPT-4o based SAMALM, the averaged SR is enhanced from 42 to 56 (a $25\%$ increase) and the averaged SS from 35 to 55 (a $36\%$ improvement). These enhancements highlight the effectiveness of incorporating a multi-critic architecture, as well as the benefits of self-verification and entropy-based score fusion mechanisms.

The ablation studies clearly demonstrate that removing the critic modules drastically impairs performance. This finding underscores the importance of our two-tier verification process, where local and global critics, coupled via an entropy-based fusion mechanism, work together to refine action proposals and ensure both robust navigation and social appropriateness. Additionally, the enhanced social scores indicate that SAMALM not only improves navigation success but also reliably maintains safe distances and facilitates smooth human interactions. Beyond these quantitative improvements, qualitative observations reveal that SAMALM generates smoother, more adaptive trajectories and is better equipped to handle dynamic changes in the environment. In practice, the critics’ feedback proves invaluable—helping the actors adjust their actions to address specific issues at both the group and individual levels, as long as these issues are effectively identified by the critics.

In summary, SAMALM’s integration of decentralized decision-making, personalized LLM-based actor modules, and dynamic self-verification mechanisms significantly advances the state-of-the-art in socially-aware navigation. This approach not only enhances the robots' ability to navigate complex environments but also ensures that their behaviors remain socially compliant, paving the way for safer and more adaptive multi-robot systems in human-populated settings. 

\vspace{-5pt}
\section{Conclusion}
\vspace{-5pt}
We first propose a novel LLM-based multi-robot social navigation approach, termed SAMALM. SAMALM represents multi-robot social norm compliance and cooperative navigation behaviors within a multi-LLM actor-critic framework. By harnessing the diverse strengths of multiple LLMs and employing rigorous self-verification mechanisms, our approach paves the way for more adaptable, resilient, and socially compliant robotic applications in complex and dynamic human environments. Moreover, the entropy-based critic evaluation fusion mechanism adaptively facilitates the incorporation of both group-level and agent-level behavior management. 


\vspace{0pt}

\typeout{}
\bibliography{main}
\bibliographystyle{IEEEtran}
\end{document}